\pdfoutput=1
\documentclass[letterpaper]{article}
\usepackage{aaai}
\usepackage{times}
\usepackage{helvet}
\usepackage{courier}
\usepackage{url}
\usepackage{graphicx}
\usepackage{color}
\DeclareGraphicsExtensions{.pdf,.png,.jpg}
\usepackage{amssymb}
\usepackage{amsmath}
\usepackage[ruled,vlined]{algorithm2e}
\usepackage{siunitx}
\usepackage{subcaption}

\definecolor{orange}{rgb}{1,0.5,0}

\newcommand{\citet}[1]{\citeauthor{#1}~\shortcite{#1}}
\newcommand{\citep}{\cite}

\newcommand{\x}{\mathbf{x}}

\title{YASENN: Explaining Neural Networks via Partitioning Activation Sequences
}

\author{
Yaroslav Zharov\thanks{\protect Contributed equally} \\
Sberbank AI Lab \\
\texttt{zharov.y.m@gmail.com}
\And  
Denis Korzhenkov\footnotemark[1] \\
Sberbank AI Lab \\
\texttt{dkorzhenkov@gmail.com}
\And
Pavel Shvechikov\footnotemark[1]\thanks{\protect Work mostly done while a researcher at Sberbank AI Lab}  \\
Samsung AI Center in Moscow \\
\texttt{shvechikov.p@gmail.com} 
\AND
Alexander Tuzhilin\\
Stern School of Business, NYU \\
\texttt{atuzhili@stern.nyu.edu}
} 

\begin{document}
\maketitle
\begin{abstract}

We introduce a novel \textit{approach} to feed-forward neural network interpretation based on partitioning the space of sequences of neuron activations. 
In line with this approach, we propose a model-specific interpretation \textit{method}, called YASENN.  
Our method inherits many advantages of model-agnostic distillation, such as an ability to focus on the particular input region and to express an explanation in terms of features different from those observed by a neural network.
Moreover, examination of distillation error makes the method applicable to the problems with low tolerance to interpretation mistakes.
Technically, YASENN distills the network with an ensemble of layer-wise gradient boosting decision trees and encodes the sequences of neuron activations with leaf indices. 
The finite number of unique codes induces a partitioning of the input space.
Each partition may be described in a variety of ways, including examination of an interpretable model (e.g. a logistic regression or a decision tree) trained to discriminate between objects of those partitions.
Our experiments provide an intuition behind the method and demonstrate revealed artifacts in neural network decision making.
\end{abstract}

\section{Introduction}
Over the last decade, the complexity of ML models increased significantly at the expense of their interpretability \cite{lipton}.
The interpretability of the model is crucial for a majority of practical applications, including validation of self-driving car steering logic, gaining insights from a generative modelling of physical system dynamics and many others.
In some domains, such as medicine and criminal justice, transparent decision making may be vitally important. 
In regulated areas, such as banking, interpretability is high-priority because of legislation compliance issues \cite{survey_guidotti}.
Therefore, a reliable interpretation method is very important in many business applications.

Interpretation of Deep Neural Networks (NN) is an active area of research,  consisting of the following main directions.
We examine each of them below together with their limitations.
\begin{itemize}
    \item Model-agnostic methods \cite{survey_guidotti}: they include (but are not limited to) those methods that distill a NN with a simple and interpretable model \cite{reallyneedbedeep,distillation2015}.
    The main limitation is that they cannot make use of the knowledge of internal mechanics of the NN.
    \item Methods explaining NNs on the basis of gradient information \cite{gradientmethods}. 
    The main limitation is that they were reported to be unstable and sensitive to irrelevant features \cite{Adebayo2018LOE,Kindermans:2017}.
    \item Case-based interpretation \cite{cbr_for_bb} is ``based on actual prior cases that can be presented to the user to provide compelling support'' for the NN decision.
    The main limitation is that they may also be of a limited utility if an object description is high dimensional (especially for tabular data) or regulation authority does not allow such an interpretation. 
    \item An interpretability could also be achieved by restricting the NN to a special interpretable design \cite{interpretable_cnn}. 
    Though appealing, this may limit the modelling power of a NN and, as a result, is of limited use for highly competitive business applications.
    \item Learning an auxiliary NN to explain an original NN is another powerful approach \cite{interpnet}.
    In our experience, however, these methods may not be transparent enough and therefore are not trustworthy. 
\end{itemize}

To address some of these limitations, we present a novel approach to the interpretation of neural networks based on partitioning the set of sequences of neuron activations.
We also developed the particular YASENN (Yet Another System Explaining Neural Networks) method that implements this approach.

The first step of YASENN consists of distilling a NN with an ensemble of gradient boosting decision trees of a special kind. 
The number of trees equals the depth of the NN, and each tree takes the neuron activations of the corresponding layer as input.
Then for each object from the dataset, we collect the sequence of leaf indices from the ensemble trees and consider the obtained sequence as a code of that object.
The key insight of our YASENN method comes from the sequential nature of both boosting and feed-forward NNs.

In the second step of the method, we describe the partitioning of the input space induced by those codes.
One of the possible descriptions can be obtained using a human-interpretable model that discriminates between objects with different codes.

One of the distinguishing features of YASENN is that it does not restrict the architecture of a NN  since it is based on distillation.
To achieve model-specificity, it makes use of the internal mechanics of a NN.
Still, another distinguishing feature of YASENN is that it has a deterministic and fast training procedure.

In addition to the basic algorithm, we also propose several powerful extensions.
Namely, the method may be enhanced to the tasks with low tolerance to interpretation mistakes.
It is also possible to provide an explanation for modified feature space or data manifold.

The contributions of this paper include:
\begin{enumerate}
    \item  We propose a novel approach to the interpretation of neural networks based on partitioning the set of sequences of neuron activations.
    \item We develop the YASENN method that implements this approach using a distilling ensemble of a special design.
    \item We demonstrate a walkthrough of applying YASENN to several datasets from different domains.
\end{enumerate}

\section{Preliminaries}
In this paper we focus only on the classification task with generalization to regression being obvious. 
\subsection{Neural Network}
We consider the task of interpreting a neural feed-forward  classifier $NN: \mathbb{R}^d \to \mathbb{R}^C$ 
trained  on the dataset $D = \{(\mathbf{x_i}, y_i)\}_{i=1}^N$, where each $\mathbf{x_i} \in \mathbb{R}^d$ and $y_i \in \{1, \dots, C\}$ are a feature vector and a class label respectively.
We also denote $\mathbf{y_i}$ for a one-hot encoding of $y_i$.
Hereinafter we will drop the subscript $i$ if it does not introduce ambiguity. 

Prediction of a NN with $M$ layers is defined as
\begin{equation}
    \begin{aligned}
    \hat{p} \left( \mathbf{y} \mid \mathbf{x} \right) 
    &=
    \rho \left(NN \left( \mathbf{x} \right) \right) \\
    &= 
    \rho ( L^{M} \circ \dots \circ L^{1} \circ L^{0} \left( \mathbf{x} \right)), 
    \end{aligned}
\end{equation}
where 
\begin{itemize}
    \item $\rho: \mathbb{R}^C \to \Delta^{C-1}$ is a map from the real-valued space to the unit simplex (usually, \verb|Softmax| function). 
    \item $L^{\ell}, ~ 1 \leq \ell \leq M,$ is a layer of a $NN$, consisting of an affine map or convolution, nonlinearity (except $L^M$) and (optionally) a regularizer(s), such as \verb|Dropout|, \verb|BatchNorm|, \verb|Pooling|, etc.
    \item $L^{0}$ is the identity function (introduced for convenience of exposition). 
\end{itemize}
Throughout this paper we treat $NN$ as a deterministic function (e.g. NN in an inference mode).    

\subsection{Streams}

For $0 \leq \ell \leq M$ define $a^{\ell}(\mathbf{x}) = L^{\ell} \circ \dots \circ L^{0}(\mathbf{x})$ to be an activation of the $\ell$-th layer on an input $\mathbf{x}$. 

A tuple of activations of a NN on an object~$\mathbf{x}$ 
\begin{equation}
    S(\mathbf{x}) =     
    \left[
        a^{0}\left(\mathbf{x}\right), 
        \dots,
        a^{M}\left(\mathbf{x}\right) 
    \right]  
\end{equation}
is referred to as a \textit{stream}  (alluding to a sequence of transformations of $\x$).

\subsection{Decision Tree}
Decision tree $T$ with $K$ leaves and $C$-dimensional output is a function 
$ T(\mathbf{x}) = W_{q(\mathbf{x})} $, 
where
\begin{itemize}
    \item $W \in \mathbb{R}^{K \times C}$ is a \textit{matrix of scores} -- $W_{ij}$ is a score for the $j$-th output of the $i$-th leaf;
    \item $q: \mathbb{R}^d \to \{1,\dots,K\}$ is an \textit{index function}, assigning a leaf for a data point.
\end{itemize}
A leaf-index function $q$ associated with a decision tree $T$ may be used to compress $\x$ into a categorical variable $q(\x)$.
Moreover, as each leaf is a connected rectangular region and the granularity of splitting depends on the variability of the target variable, the decision tree respects both \emph{input space} and \emph{target proximities}.

\subsection{Distillation}
Distillation assumes that one tries to use the output of the original complicated black-box (``teacher'') as a so-called soft target for another model (``student''), typically more simple, fast or interpretable and less cumbersome.
In other words, one treats the given teacher model just as a deterministic function and optimizes the parameters of a student model to approach that function as precisely as possible on the specified transfer dataset.

One notable property of distillation is the immunity to overfitting to noise \cite{towardsrobustness}.
This is the case because, while distilling, one tries to reproduce the original deterministic contour lines with the model of another kind.

\section{Method} 
Case-based reasoning and prototypes learning partition the input space into segments and assign a representative to each segment \cite{prototype_selection,bcm}. 
In this paper, we follow a similar idea of segmentation.
However, we do not select a strict representative for each segment. 
Instead, we propose to divide the input space into regions of low variation in NN prediction and find \emph{interpretable descriptions} for these regions. 
To increase interpretation transparency, we propose to consider the final partitioning of the input space as clustering that preserves proximities (closeness) between objects both in the \emph{prediction} and the \emph{input} spaces. 

To explain the decisions of a NN, we next examine the process of decision making, including the transformation of activations, before describing our YASENN method itself.
Note that studying activations is a long-standing research topic. 
In particular, \cite{VisConvNet} showed that neurons in convolutional NN activate on colours and geometric shapes that are interpretable for a human.
Activations were also examined in a context of generalization \cite{SingleDirGener,InsightCanonCorr}, perceptual metrics \cite{DeepFeatures},   style transfer \cite{ArtStyle} and representation similarities \cite{InsightCanonCorr}.  

We base our interpretation method on the notion of a \textit{stream}, that captures the essential information about NN decision making by manifesting dependencies hidden in NN parameters. 
The concept of a stream was described above. 
A closely related concept was successfully used in \cite{DeepkNN,interpnet,neurorule}.

Note that streams are difficult to work with directly due to the high dimensionality, continuity and absence of natural distance function capturing the geometry of this space. 
However, since the space of streams, deterministic transformations of the input space, is structured, its intrinsic dimension cannot be more than that of the input space. 
Therefore we maintain that streams can be compressed to a lower dimensional space, and that space can be subsequently partitioned and interpreted.
To achieve this, we define the two steps of our method: compression and inspection, that are described below.

\subsection{Step One: Compression}
To utilize all the activations we propose to fit a distilling ensemble of trees $\langle T^{0}, \dots, T^{M-1} \rangle$ -- a separate tree for each but final layer of a NN.
In particular, the $\ell$-th tree takes $a^{\ell}\left(\x\right)$ as input and distills the part of the network from $\ell$-th layer onward.
The trees are linked in a way similar to boosting due to a heavily sequential nature of both models  -- the following module (tree or layer) learns such dependencies that were not yet captured with the previous ones.

We train the ensemble in gradient boosting manner with MSE loss function and raw logits $NN \left(\x\right)$ as a target variable with the only distinction from \cite{gradboost}:
each tree $T^\ell$ operates on the space of activations of the corresponding layer, not on the random subspace of features
(see Algorithm \ref{alg:boostedtree}).

We define a \textit{discretized stream} of $\x$, $DS(\x)$, as a tuple of leaf indices of each tree in a distilling ensemble $\langle T^{0}, \dots, T^{M-1} \rangle$ :
\begin{equation}
    DS(\x)  
    = 
    \left[ 
        q^0 \left(a^0 \left(\x \right)\right), \dots, 
        q^{M-1} \left( a^{M-1} \left( \x \right) \right)
    \right] 
\end{equation}

A discretized stream $DS(\x)$ is actually a compressed stream $S(\x)$ with reduced both dimensionality and cardinality.

The set of unique discretized streams $\{DS(\x)   \mid \x\in~D \}$ can also be enumerated in arbitrary order (e.g. lexicographic) and each $\x$ is assigned a \textit{stream label} -- an index of a $DS(\x)$ in the order.

\begin{algorithm}[ht]
    \DontPrintSemicolon
    \SetKwInOut{Input}{input}
    \SetKwInOut{Output}{output}
    \SetKwInOut{Variables}{variables}
    \SetInd{0.em}{.5em}
    \caption{Fitting the distilling ensemble}
    \label{alg:boostedtree}
    \Input{
        $D=\{(\mathbf{x_i}, y_i)\}_{i=1}^N$ (dataset);\\
        $NN$ (neural network)
    }
    \Output{
        $T^{0} \dots T^{M-1}$ (layer-wise boosting trees) 
    }
        
    \Variables{$\hat{Y}, G$ (matrices of size $N \times C$)}
    \BlankLine
    \Begin{
        \For(\tcp*[f]{Constant prediction}){$i \gets 1 \dots N$}{
            $\hat{Y}_{i,*} \gets \frac{1}{N} \sum\limits_{j=1}^N{NN(\mathbf{x_j})}$
        } 
        \For{$\ell \gets 0  \dots M-1$}{
        
            \For{$i \gets 1 \dots N$}{
                $G_{i,*} \gets NN(\mathbf{x_i}) - \hat{Y}_{i,*}$
                \tcp*[r]{Gradient}
                
                $A_{i,*} \gets a^{\ell}(\mathbf{x_i})$
                \tcp*[r]{Activation}
                
            }
            \BlankLine
            \tcp{
                Fit the tree 
            }
        
            $T^{\ell}$ $\gets$ {$DecisionTreeRegressor(A, G)$}\;
            
            
            \BlankLine
            \tcp{Update prediction}
            $\hat{Y} \gets \hat{Y} - T^{\ell}(A)$
        }
    }
\end{algorithm}

\subsection{Step Two: Inspection}

To understand the logic behind a particular partitioning we need a second step -- inspection, which aims at identification of reasons behind the differences among discretized streams. 
Restriction of NN to be deterministic makes it possible to attribute such differences only to the input space, that is to differences in~$\x$.

For the particular method used at this step, we refer to as the \textit{inspector}.

For example, we can treat the stream label of $\x \in \mathbb{R}^d$ as its class label and solve a multiclass classification problem with a decision tree as an inspector.  
Paths from the root of the fitted tree to its leaves will signify the rules discriminating stream labels in terms of object features.
However, this approach may be impractical for a large number of stream labels. 
Inspector may vary depending on the task at hand or legislation requirements. 
In general, any method that provides an interpretable description of objects of the same stream label suffices.  
In this sense, possible inspectors include,  but are not limited to 
\begin{enumerate}
    \item Interpretable discriminative models (e.g. logistic regression, decision tree, rule mining)
    \item Exploratory techniques (e.g. averaging object features for each stream label, describing each region with a prototype, etc.). 
\end{enumerate}

Depending on the particular choice of the model, we may use it in either of ways:
\begin{enumerate}
    \item \textit{Self-descriptive} (e.g. feature averaging) -- describes the stream label without contrasting it with any other objects. 
    \item \textit{Contrastive} (e.g. logistic regression) -- discriminates between objects of the stream label and some other group of objects.
\end{enumerate}

The choice of the contrastive group (i.e. the negative class for discriminative modelling) is reminiscent of the baseline choice from gradient interpretation methods \cite{gradientmethods}. 

\subsection{Discussion}
Here we justify the design of our compression step.

\subsubsection{Why an Ensemble?}
One could suggest to distill a neural network with just a single decision tree assuming the stream $S(\x)$ as an input for the tree and to compress the stream by applying the leaf-index function.

But this naive approach may backfire: while first few activations are most interpretable since they are primitive transformations of the original feature vector; 
last activations contain much more information about the prediction and, finally, activation $a^M$ is the distillation target itself.
A single distilling tree, thus, may exploit only the last activations (even if we remove $a^M (\x)$ from input), since they are most associated with NN prediction. 

\subsubsection{Why Linked Trees?}
While usage of independent layer-wise trees is the straight-forward solution, it may cause some of the trees to learn similar dependencies. 
To utilize the limited number of trees more efficiently, we need to reduce this redundancy.

\subsubsection{Why Logits?}
While it is possible to train the ensemble with the prediction $\widehat p(\mathbf{y}| \mathbf{x})$ as a target, we note that it is more preferable to use raw logits $NN(\x)$ as a target for the following reasons: 
\begin{enumerate}
    \item Aggregation operation (e.g. \verb|Softmax|) is not bijective, making it possible to lose information about equivalent (up to an additive constant) outputs.
    \item Exponentiation, which is an integral part of the usual \verb|Softmax|, requires many deep trees for satisfactory distillation.
\end{enumerate}
In addition, distilling raw logits instead of probability prediction drops the necessity to adapt the procedure to classification or regression. 

\section{Extensions}
In this section, we explore some properties of YASENN and describe several ways to enhance the proposed method.

\subsection{Input Space Modification} 
The inspector treats input space flexibly. 
To obtain the description, the input space can be modified to increase clarity. 
Instead of the input space seen by a NN one can feed in the space of aggregated metrics. 
For example, this could be the segmentation of images instead of raw pixels or aggregated metrics for tabular data (i.e. condensing the group of features into lower dimensional decorrelated set or quantiles). 

Model-specific methods often lose this property (e.g. those relying on gradient information). 
YASENN, while being model-specific, preserves the ability to work with modified input spaces. 

\subsection{Adaptive Explanation} 
Alongside with input space change -- input manifold restriction is also possible.
Examples of restricting the input manifold may include narrowing the customer set (e.g. young and poor) as well as explaining adversarial examples for a particular NN  \cite{adversarial}.

When operating with lack of data near the object of interest we could use the procedure of adaptive explanation: consequently shrink input manifold closer to the region of interest, sample new objects to this region and refit the distilling ensemble in this area.

\subsection{Deliberate Interpretation} 

In practice, it is sometimes more important to explain most decisions extremely well than to make a uniformly modest explanation. 

There are two potential problems that may cause the deliberately unreliable interpretation.
We propose to use \emph{adaptive explanation} to eliminate them.
However, if unavailable, one can use criteria, described below.
\subsubsection{High distillation error}
Select the threshold of distillation error guided by either absolute value of the error or a fraction $\alpha$ of the training set size.
Remove the objects with the error violating that threshold from the training set of any inspector.
Avoid providing interpretation for new objects of that kind.

Even if the compression is trained to minimize MSE, one can be more concerned with the distillation performance in the probability space. 
For this reason, we recommend to apply the $\rho(\cdot)$ (think of \verb|Softmax|) to ensemble predictions and estimate the discrepancy in probability domain (e.g. with a cross-entropy loss).

\subsubsection{Underpopulated discretized streams} 
Select the low-populated discretized streams guided by either a threshold of minimal population per label or a fraction $\gamma$ of the training set size.
Avoid to train an inspector for such streams and to explain NN decision for their objects. 

\subsection{Anomaly Detection}
The majority of possible discretized streams are not populated because of the nonrandom nature of the tree splits.
It is often the case that conditionally on the previous split history an entropy of partitioning objects into leaves is far from minimum.

That property could be used for rough anomaly detection. 
In particular, if the discretized stream of an object of interest matches no discretized stream from the training set, chances are this object is anomalous and requires additional attention and/or special treatment. 
However, the converse does not hold in general: an object corresponding to a densely populated stream label can also be an anomaly.

\subsection{Improved Ensemble}
For the sake of better generalization we recommend to extend Algorithm \ref{alg:boostedtree} by fitting a multiplier $\beta_\ell$ for each tree $T^\ell$ as proposed in \cite{gbdt_with_multiplier}.

If the combinatorial growth of the number of possible discretized streams is unwanted, it can be reduced by applying ensemble trees of increased complexity, e.g. oblique trees \cite{multivariate_dt,oblique_dt} or SVM trees \cite{svm_dt}.
Oblique trees are especially relevant since a linear combination of activations in the nodes of the decision tree can additionally learn the decorrelation structure (similar to PCA).
This linear combination may also depend on the previous split history, which is a lot more flexible then a simple PCA fitted once.

\section{Experiments}
We apply YASENN to various feed-forward NN architectures trained on datasets from different domains. 
In our experiments we would like to:
\begin{enumerate}
    \item showcase the application of \textit{deliberate interpretation} and \textit{adaptive explanation}
    \item show that the objects with different stream labels are less similar than those with the same label
    \item demonstrate that discretized streams could be used to provide some useful insights about the NN decision-making process.
\end{enumerate}

Several possible metrics of interpretation quality have been proposed recently, but we cannot measure the quality of our interpretation with them for the following reasons.

\begin{itemize}
    \item Performance of distilling ensemble is a part of the interpretability-accuracy trade-off driven by the boosting trees hyperparameters on the one hand and limited human attention budget \cite{lime} on the number of streams on the other hand.
    We did not search for the optimal point of that trade-off and left it for further research.
    However, due to the greedy procedure of fitting decision trees, the ensembles presented in this section provide relevant partitioning of the input space despite they may be not the best ensembles of given complexity in terms of \emph{accuracy} and \emph{fidelity}. 
    \item Performance of the discriminative inspectors denotes the trustworthiness of the explanation provided by that inspector. 
    However, the main target of this paper was to develop a method of stream space partitioning, and, therefore selection of the best kind of inspectors was out of scope.
    A researcher can select the best inspector according to \emph{unambiguity}, \emph{complexity} and \emph{input shift invariance} proposed in  \cite{beta,Kindermans:2017}.
    We report the quality of inspectors in terms of \emph{ROC AUC} to demonstrate the reliability of explanation. 
\end{itemize}

None of the existing interpretation approaches can be directly compared with YASENN: 
It provides explanations somewhere between the local and the global scopes, describing stream in general, which is not a common case.
Also, we cannot compare the proposed very different (but connected) steps of our procedure individually with independent baselines.

All NNs in our experiments were implemented with PyTorch \cite{pytorch} and trained with Adam optimizer \citep{adam}, learning rate \num{3e-4}.
To fit the distilling ensemble we used Scikit-learn \cite{scikit-learn} implementation of CART tree \cite{cart84-2}.

\subsection{Gaussian Mixture}
To provide intuition behind the compression step we consider a simple problem for which discretized streams can be visualized.

\begin{figure*}[!t]
    \centering
    \begin{subfigure}[t]{.3\linewidth}
        \centering
        \includegraphics[height=\linewidth]{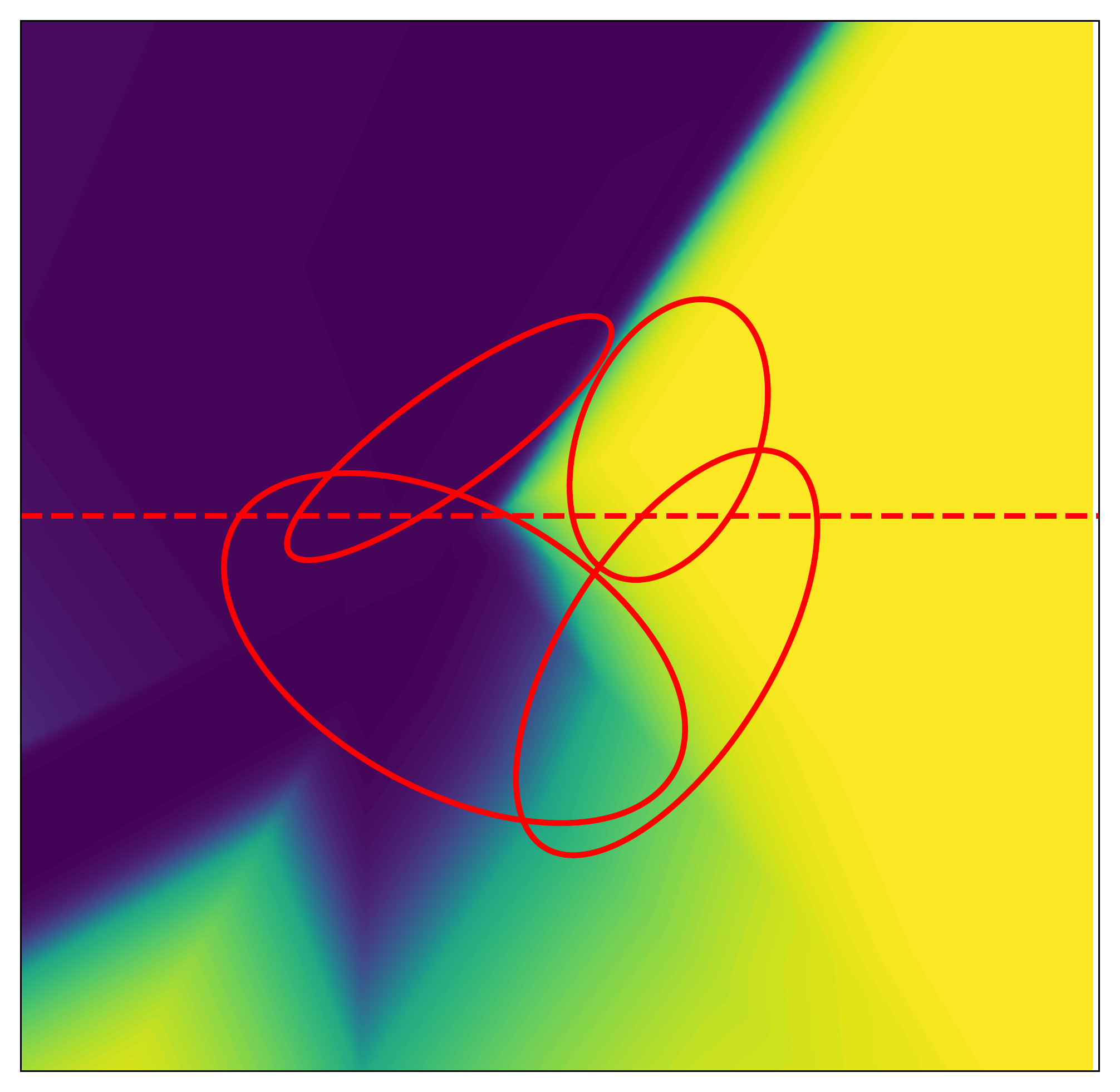}
        \caption{}
    \end{subfigure}
    ~
    \begin{subfigure}[t]{.3\linewidth}
        \centering
        \includegraphics[height=\linewidth]{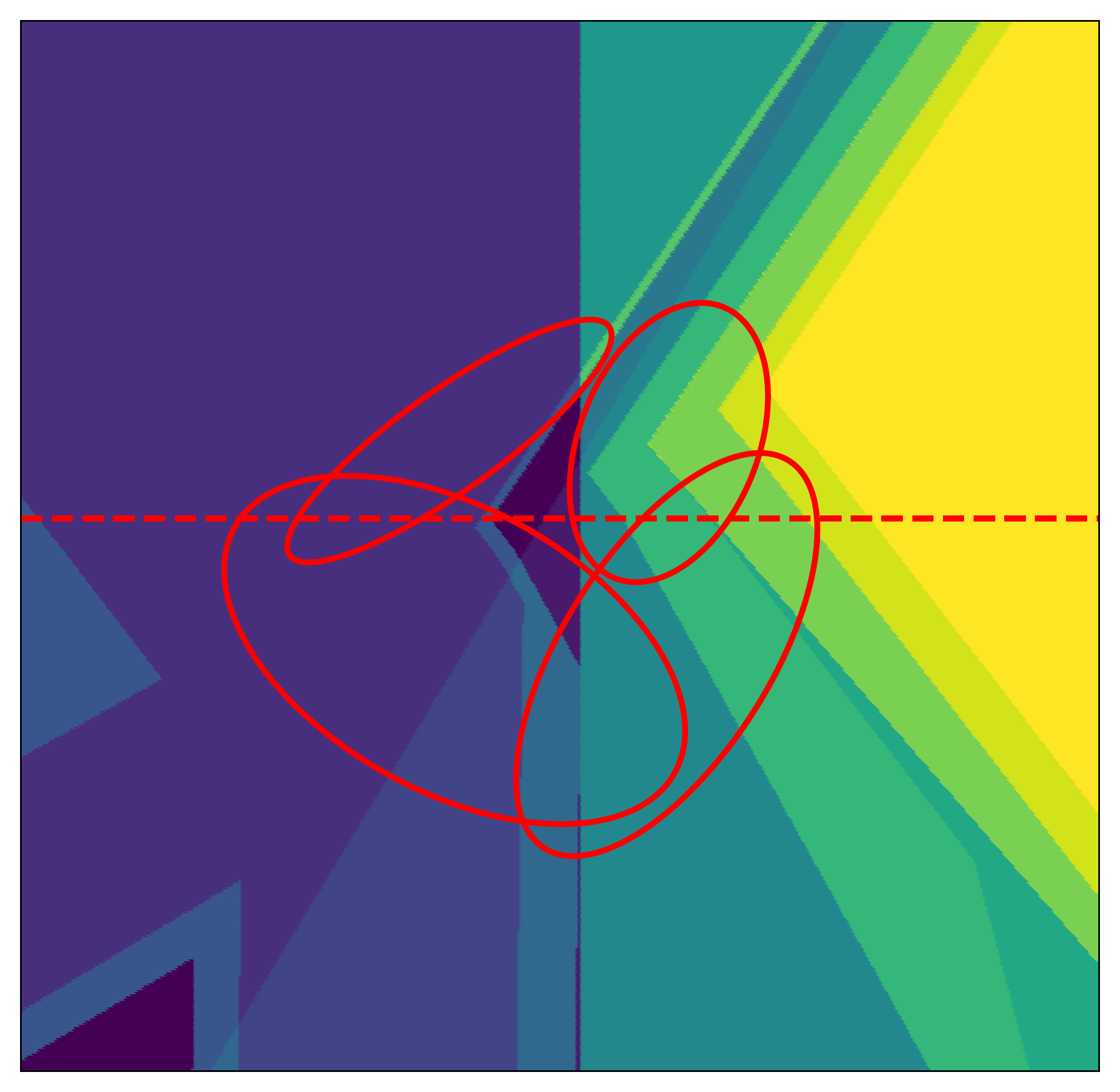}
        \caption{}
    \end{subfigure}
    ~
    \begin{subfigure}[t]{.3\linewidth}
        \centering
        \includegraphics[height=\linewidth]{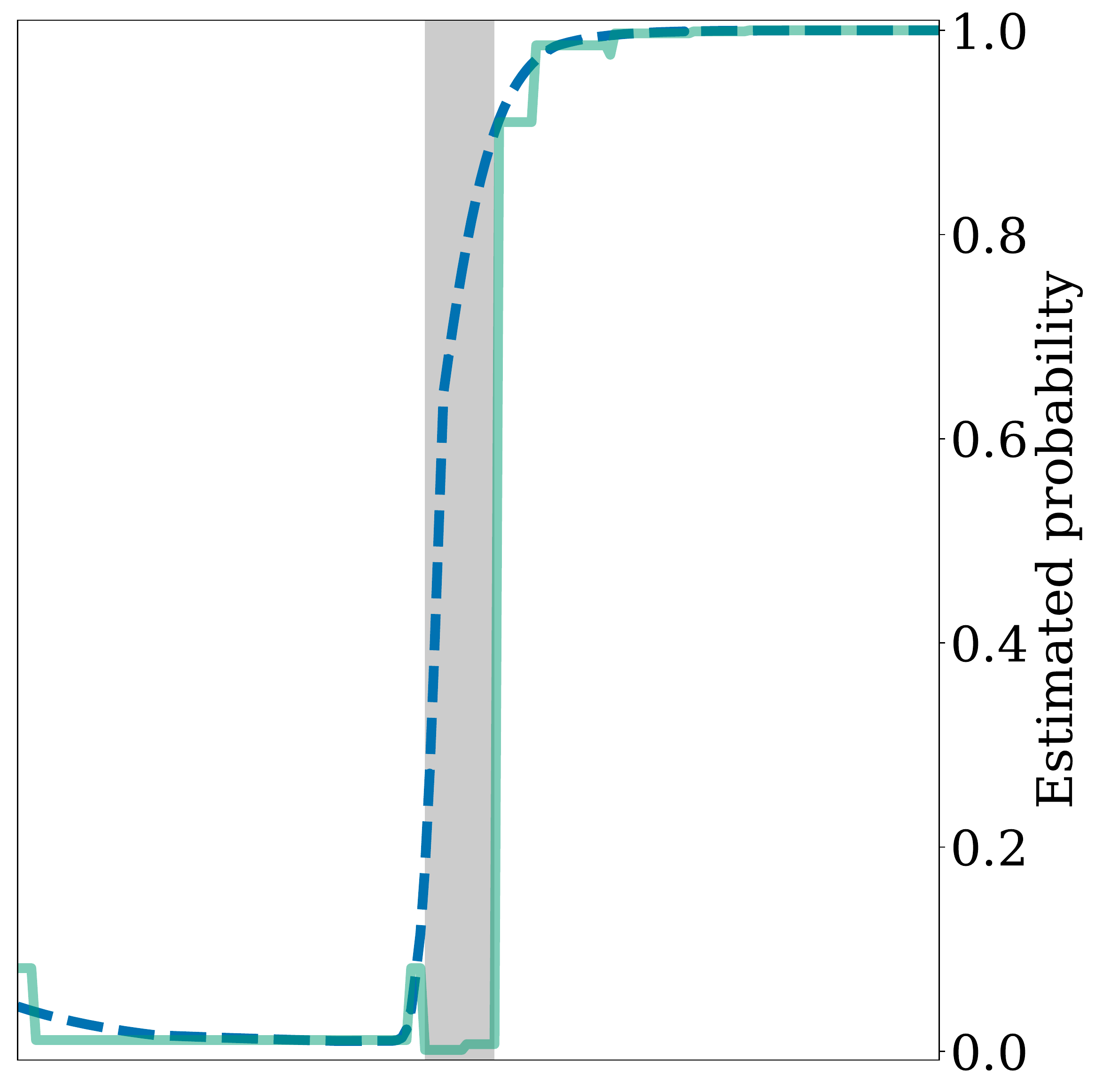}
        \caption{}
    \end{subfigure}
    \caption{
    Data was sampled from Gaussian mixture model and divided into two classes.
    `2-sigma' ellipses of mixture components are shown.
    The horizontal dashed line represents a cut to be analyzed.
    (a) The probability of class 1 estimated with the NN.
    The more light the colour is, the higher the estimation is.
    (b) Points of the same colour have the same stream label.
    (c) A solid line represents the probability of class 1 along the profile cut estimated by the NN, a dashed one -- by the distilling ensemble.
    Filled area marks the region where the cross-entropy between the predicted class distributions is above the threshold selected according to the training sample.
    This area mostly corresponds to the low-density region of the space.
    }
\label{fig:gaussian_mixture}
\end{figure*}

\subsubsection{Setting}
We sampled 7,500 objects from four Gaussians and divided them into two classes. 
Negative class was assigned to the objects coming from left components and positive class -- to those from the right components (Figure \ref{fig:gaussian_mixture}(a)). 
We also flipped class of 2\% of randomly chosen objects. 

A fully connected neural network 
[\verb|Linear|$(2, 2)$ - \verb|PReLU|$()$](5 times) - \verb|Linear|$(2, 2)$ was trained for 1,000 epochs to solve this classification problem.

The depth of ensemble trees was set to one for layers 1-3 and to two for layers 4-5. 
We increased the depth of the last trees to simplify figures while preserving all the details.

\subsubsection{Findings}
The network learned a complicated decision boundary with artifacts at the bottom-left corner (Figure \ref{fig:gaussian_mixture}(a)). 

The compression partitioned the space into regions of the same color (Figure \ref{fig:gaussian_mixture}(b)). 
Each colour corresponds to a discretized stream and is proportional to the probability predicted by the compression.

Some notable findings from Figure \ref{fig:gaussian_mixture} (a,b) include the following.
\begin{itemize}
    \item  Area of the same stream label may include disconnected regions due to splits on high-level interactions of input features. 
    \item The ensemble partitioned the space more frequently along the decision boundary, because of the high gradient of NN outputs with respect to its inputs in these areas. 
\end{itemize}

The error between predictions of the NN and the ensemble was the highest in the regions of low data density -- in the center and in the bottom-left corner (dark triangle regions in Figure \ref{fig:gaussian_mixture}(b)). 
To illustrate this we plot probability predictions (Figure \ref{fig:gaussian_mixture}(c)) along the dashed horizontal cut from Figure \ref{fig:gaussian_mixture}(a,b). 

Guided by the procedure of \textit{deliberate interpretation}, we identified the threshold isolating the $5\%$ of the biggest distillation errors on the training set. 
According to this threshold, the filled region in Figure \ref{fig:gaussian_mixture}(c) corresponds to uncertain distillation and, thus, to unreliable explanation. 
We recommend using \textit{adaptive explanation} procedure to refine the distillation in this region. 

\subsection{MNIST}
We illustrate the applicability of our method to convolutional NNs on the MNIST \cite{mnist} classification problem. Experiments with MNIST include two settings described below.

\subsubsection{Setting 1}
A convolutional network 
\verb|Conv2d|$(1, 16, 3)$ - \verb|ReLU|$()$ - 
\verb|Conv2d|$(16, 16, 3)$ - \verb|ReLU|$()$ - 
\verb|Conv2d|$(16, 16, 3)$ - \verb|ReLU|$()$ - 
\verb|Flatten|$()$ - \verb|Linear|$(16 \times 22 \times 22, 10)$
was trained for 5 epochs. 

To illustrate \textit{adaptive explanation} we fitted the distilling ensemble only on objects classified by NN as 7. 
The depth of ensemble trees was set to two. 
The minimum number of samples for each leaf of each tree was set to 2\% of the ensemble training set.

We used a logistic regression as a one-vs-all inspector.
For each discretized stream we fitted the inspector to distinguish between its objects and all other MNIST objects.
Objects exceeding the threshold of 5\% cross-entropy distillation discrepancy were excluded from the ``one'' component.

\begin{figure}[htbp]
        \centering
        \includegraphics[width=.8\linewidth]{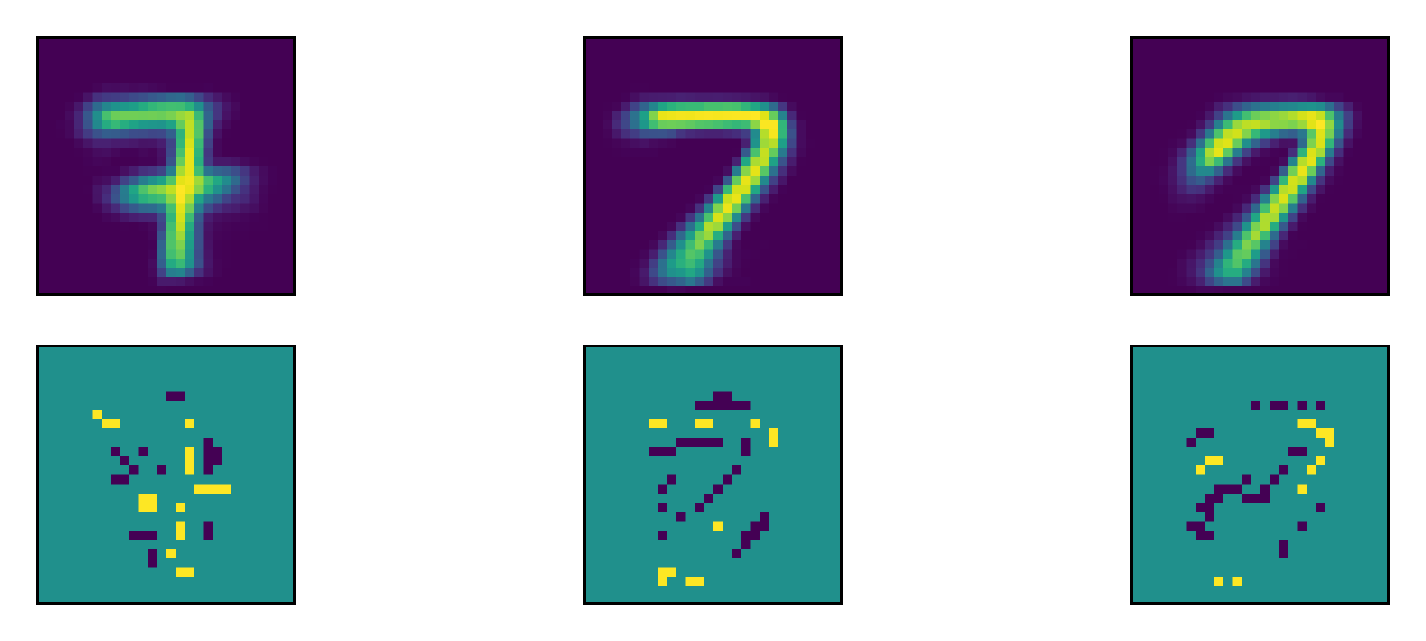}
        \caption{Three discretized streams that lead to the high probability of class 7 and signs of coefficients from respective inspectors (top row). The coefficients related to light-colored pixels have a positive sign and to dark-colored pixels -- a negative one (bottom row).}
        \label{fig:mnist}
\end{figure}

\subsubsection{Findings 1}
The averaged images of three most indicative discretized streams are shown in the top row of Figure \ref{fig:mnist}. 
They have a relatively high population (at least one hundred samples).
One may observe that different discretized streams refer to different digit appearances.
The pixels that are important for the fitted inspectors (bottom row of Figure \ref{fig:mnist}) are human-interpretable.
Since the inspectors have ROC AUC not less than 0.996, the explanation is satisfactory.
\subsubsection{Setting 2}
The same NN as in the previous setting was used.
To illustrate the exploration of frequent misleading patterns we fitted a distilling ensemble on the objects misclassified by NN. 
The depth of ensemble trees was set to 1, 2, 2 and 2 respectively, with minimum impurity decrease set to 1, 0.9, 0.9 and 0.9.

\subsubsection{Findings 2}

\begin{figure}[ht]
    \centering
    \begin{subfigure}[t]{.3\linewidth}
        \centering
        \includegraphics[height=.7\linewidth]{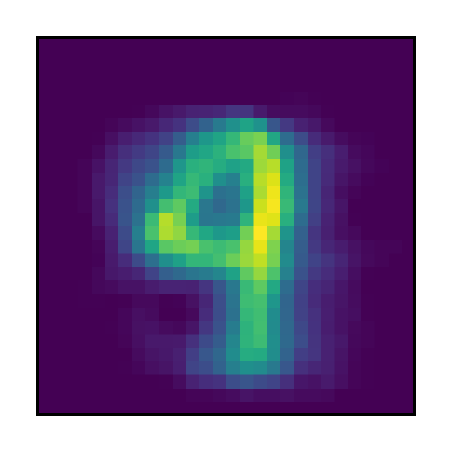}
        \caption{averaged picture}
        \label{fig:mnist_4:avg}
    \end{subfigure}
    ~
    \begin{subfigure}[t]{.3\linewidth}
        \centering
        \includegraphics[height=.7\linewidth]{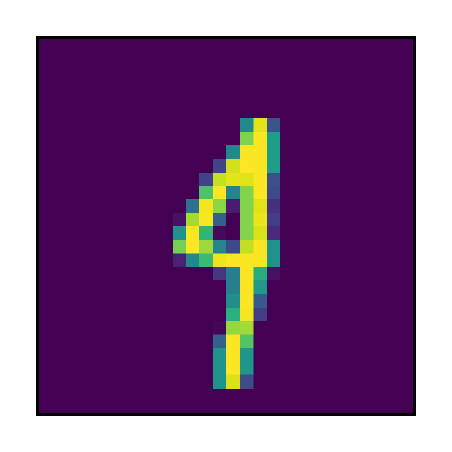}
        \caption{misleading 4}
        \label{fig:mnist_4:misleading_4}
    \end{subfigure}
    ~
    \begin{subfigure}[t]{.3\linewidth}
        \centering
        \includegraphics[height=.7\linewidth]{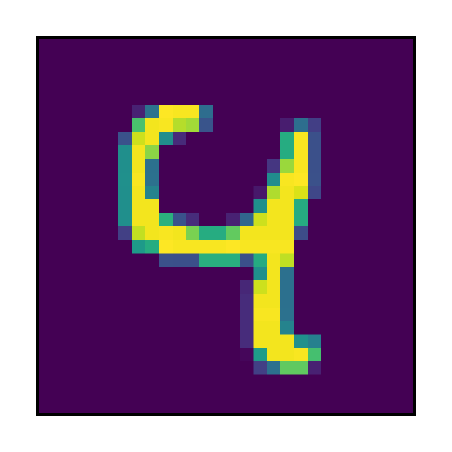}
        \caption{misleading 9}
        \label{fig:mnist_4:misleading_9}
    \end{subfigure}
    \caption{Showcase of misleading discretized stream containing objects that look like digits 4 and 9 at the same time.}
    \label{fig:mnist_4}
\end{figure}

Despite the NN performs well (test-time accuracy $98.1\%$) and there is no much misclassified data, we have found some persistent patterns.
One of the discretized streams, shown in Figure \ref{fig:mnist_4}, contains objects, which look like digits 4 and 9 at the same time.
Figure \ref{fig:mnist_4}(b) depicts an image which has probabilities for classes 9 and 4 of $0.44$ and $0.42$ respectively (as predicted by NN).
Figure \ref{fig:mnist_4}(c) depicts an image which has those probabilities equal to $0.01$ and $0.98$ respectively, while the ground-truth label is 9.

\subsection{IMDB}
Following \citep{learningtoexplain} we verify our method on the sentiment classification problem on IMDB dataset \cite{imdb}. 
Here we demonstrate that YASENN produces meaningful insights about the nature of network streams.
\subsubsection{Setting}
The dataset contains bag-of-words representations of movie reviews.
The rare (term frequency $< 50$) and frequent (term frequency $> 10,000$) features were deleted through the preprocessing step.
The resulting number of features is about 6,000 while the training data itself consists of 12,500 positive (class 1) and 12,500 negative (class 0) examples.

A feed-forward network 
\verb|Linear|$(6192, 500)$ - \verb|Dropout|$(0.9)$ - \verb|LeakyReLU|$()$ - 
\verb|Linear|$(500, 50)$ - \verb|Dropout|$(0.9)$ - \verb|LeakyReLU|$()$ - 
\verb|Linear|$(50, 1)$ was trained for 10 epochs with batch size 64.

Layer-wise trees were fitted with the depth set to 1, 1 and 2 for each of the trees respectively for the purposes of good stream quality.

We used a $L_1$-regularized logistic regression as an inspector for this experiment in all the cases described below.

\begin{figure}[ht]
    \centering
    \begin{subfigure}[t]{.45\linewidth}
        \centering
        \includegraphics[height=.75\linewidth]{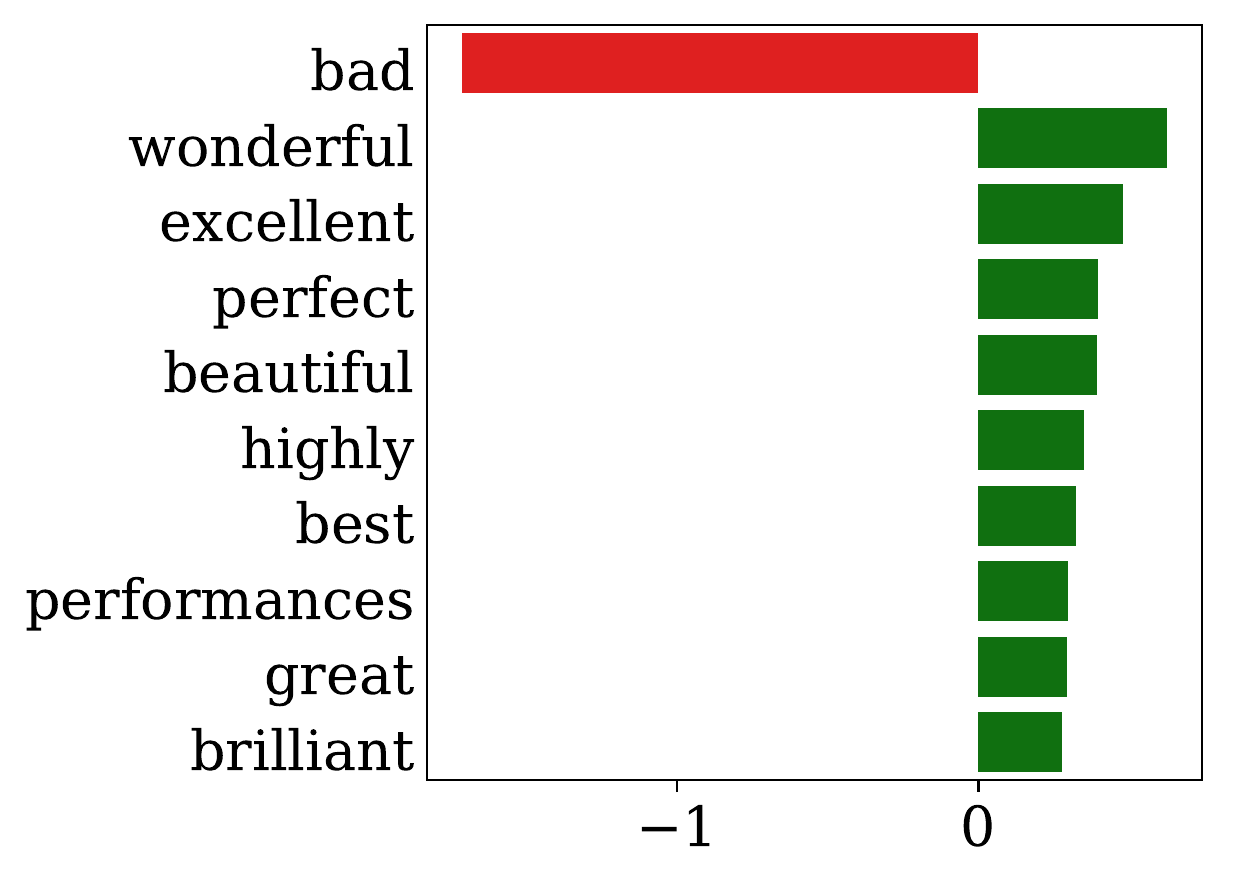}
        \caption{Presence of positive words}
    \end{subfigure}
    ~
    \begin{subfigure}[t]{.45\linewidth}
        \centering
        \includegraphics[height=.75\linewidth]{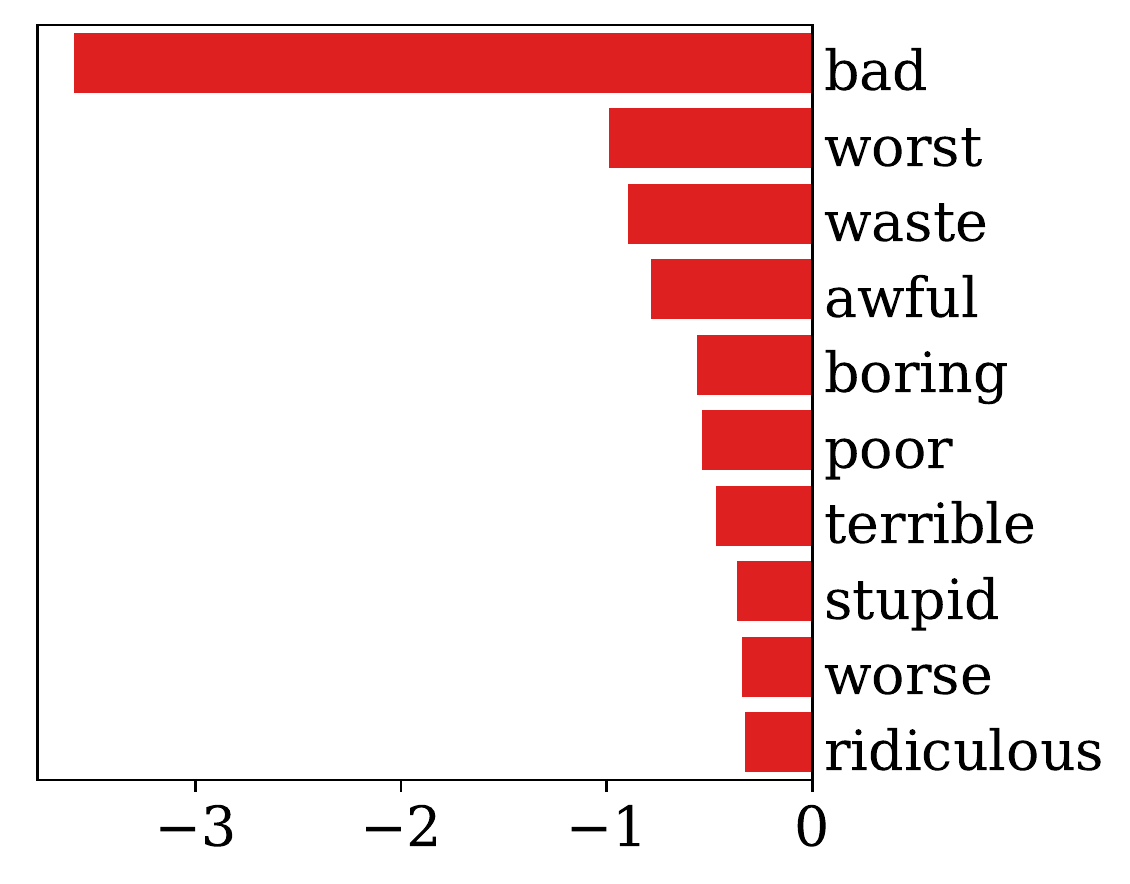}
        \caption{Absence of negative words}
    \end{subfigure}
    \caption{Features with top-10 highest absolute value of coefficients from one-vs-all logistic regressions for two discretized streams leading to a high probability of the positive class.}
    \label{fig:imdb}
\end{figure}

\subsubsection{Findings}
We generated 16 discretized streams.
To illustrate how different they can be in spite of close predictions, we selected 2 streams which had the ensemble prediction of the positive class close to 1.
NN and the ensemble predict the same major class for 100\% and 94\% of training objects for the first and second discretized streams respectively.
Population of each stream was more than 1,000.

To make intuition of which words characterize them, two one-vs-all inspectors were fitted for these streams (with ROC AUC 0.95 and 0.89 respectively), and we checked which features (i.e. words) were the most important for their respective inspectors.

Figure \ref{fig:imdb} presents 10 of the most important words for the selected streams.
Note, that the absence of word `bad' is the most deciding factor for both streams, while other important words are different not only by coefficient value but also by intent.
The presence of positive words led to a high probability of the first selected discretized stream (a), and the absence of negative words led to a high probability of the second stream (b).
It is a human-interpretable difference between them: 
the first stream contains explicitly positive reviews and the second one includes reviews that just do not have negative words.

This provides an insight that the network decision in the second stream may not be completely trustworthy since the NN relies mostly on the absence of negative terms.

\section{Related Works}

This section provides the connection between YASENN and existing interpretation methods.

Our method is closely connected to TREPAN \cite{trepan}, which extracts rules from NNs with a special kind of decision trees. 
Like this method, we also use trees to interpret the decision-making procedure. 
Unlike it, we use the internal structure of a NN to gain more knowledge.

\cite{LORE} proposed LORE method for local rule extraction. 
LORE uses a special procedure to sample more data near the object of interest, which looks similar to our adaptive explanation extension.

Like \cite{distilltransparent,distilltransparentaudit,icu} we distill a NN with a complex unexplainable model.
But unlike it, we interpret the extracted partitioning instead of examining the student model.

Like a prototype classifier network \cite{prototypes} we partition the input space with respect to the intrinsic decision-making process.
However, in general, YASENN does not return a prototype for each discretized stream.
One can produce it in a way appropriate for the application area. 
Also, we do not restrict the NN to a special architecture.

The most relevant papers for our method are those that deal with NN activations.
Here we describe only the crucial discrepancies between them and YASENN.

Unlike NeuroRule \cite{neurorule}, we work with activations in a more general way: we take into account the whole layer rather than process neurons one by one.
In addition, YASENN was designed for deep NNs, and it gains benefits of their depth.

Unlike InterpNET \cite{interpnet}, we explore activations in a layer-wise manner instead of considering them all at once.

Unlike DeepkNN \cite{DeepkNN}, which is the closest in spirit to our approach, we preserve the sequential essence of neuron activations and do not make any assumptions about the distance function in the layer activations space.

\section{Conclusion and Future Work}
We have presented YASENN method that incorporates the approach for interpreting NNs, proposed in the paper.
This method has the following benefits:
\begin{enumerate}
    \item it derives from distillation the applicability to the modified input manifold and the resistance to overfitting to noise
    \item our method is not constrained to a single type of inspector
    \item YASENN gains information from intrinsic network transformations
    \item deliberate interpretation is available for the areas which require strong control
    \item the proposed method uses decision trees in the compression step and therefore is deterministic and has low computational complexity.
\end{enumerate}
We have empirically tested YASENN on several diverse data types and showcased its ability to find useful information about the NN.

Although it has nice features, YASENN also has the following limitations.
First, it is sensitive to a combinatorial growth of the number of possible discretized streams.
The second problem is the determination of boosting tree hyperparameters.

As a future research, we plan to work on how to apply YASENN to recurrent neural networks.
We also plan to enhance YASENN with the usage of gradient information and more powerful tree-based algorithms.
Finally, we are working to equip our method with certain distance functions in the space of discretized streams.

\fontsize{9.0pt}{10.0pt}
\selectfont
\bibliographystyle{aaai}
\bibliography{references}

\begin{thebibliography}{}

\bibitem[\protect\citeauthoryear{Adebayo \bgroup et al\mbox.\egroup
  }{2018}]{Adebayo2018LOE}
Adebayo, J.; Gilmer, J.; Goodfellow, I.; and Kim, B.
\newblock 2018.
\newblock Local explanation methods for deep neural networks lack sensitivity
  to parameter values.

\bibitem[\protect\citeauthoryear{Ancona \bgroup et al\mbox.\egroup
  }{2018}]{gradientmethods}
Ancona, M.; Ceolini, E.; {\"O}ztireli, C.; and Gross, M.
\newblock 2018.
\newblock Towards better understanding of gradient-based attribution methods
  for deep neural networks.
\newblock In {\em International Conference on Learning Representations}.

\bibitem[\protect\citeauthoryear{Ba and Caruana}{2014}]{reallyneedbedeep}
Ba, J., and Caruana, R.
\newblock 2014.
\newblock Do deep nets really need to be deep?
\newblock In Ghahramani, Z.; Welling, M.; Cortes, C.; Lawrence, N.~D.; and
  Weinberger, K.~Q., eds., {\em Advances in Neural Information Processing
  Systems 27}. Curran Associates, Inc.
\newblock  2654--2662.

\bibitem[\protect\citeauthoryear{{Barratt}}{2017}]{interpnet}
{Barratt}, S.
\newblock 2017.
\newblock {InterpNET: Neural Introspection for Interpretable Deep Learning}.
\newblock {\em ArXiv e-prints}.

\bibitem[\protect\citeauthoryear{Bien and
  Tibshirani}{2011}]{prototype_selection}
Bien, J., and Tibshirani, R.
\newblock 2011.
\newblock Prototype selection for interpretable classification.
\newblock {\em Ann. Appl. Stat.} 5(4):2403--2424.

\bibitem[\protect\citeauthoryear{{Breiman} \bgroup et al\mbox.\egroup
  }{1984}]{cart84-2}
{Breiman}, L.; {Friedman}, J.~H.; {Olshen}, R.~A.; and {Stone}, C.~J.
\newblock 1984.
\newblock {\em Classification and Regression Trees}.
\newblock Statistics/Probability Series. Belmont, California, U.S.A.: Wadsworth
  Publishing Company.

\bibitem[\protect\citeauthoryear{Brodley and Utgoff}{1995}]{multivariate_dt}
Brodley, C.~E., and Utgoff, P.~E.
\newblock 1995.
\newblock Multivariate decision trees.
\newblock {\em Machine Learning} 19(1):45--77.

\bibitem[\protect\citeauthoryear{Carlini and Wagner}{2017}]{towardsrobustness}
Carlini, N., and Wagner, D.
\newblock 2017.
\newblock Towards evaluating the robustness of neural networks.
\newblock In {\em 2017 IEEE Symposium on Security and Privacy (SP)},  39--57.

\bibitem[\protect\citeauthoryear{Che \bgroup et al\mbox.\egroup }{2016}]{icu}
Che, Z.; Purushotham, S.; Khemani, R.; and Liu, Y.
\newblock 2016.
\newblock {{I}nterpretable {D}eep {M}odels for {I}{C}{U} {O}utcome
  {P}rediction}.
\newblock {\em AMIA Annu Symp Proc} 2016:371--380.

\bibitem[\protect\citeauthoryear{Chen \bgroup et al\mbox.\egroup
  }{2018}]{learningtoexplain}
Chen, J.; Song, L.; Wainwright, M.; and Jordan, M.
\newblock 2018.
\newblock Learning to explain: An information-theoretic perspective on model
  interpretation.
\newblock In Dy, J., and Krause, A., eds., {\em Proceedings of the 35th
  International Conference on Machine Learning}, volume~80 of {\em Proceedings
  of Machine Learning Research},  883--892.
\newblock Stockholmsm{\"a}ssan, Stockholm Sweden: PMLR.

\bibitem[\protect\citeauthoryear{Craven and Shavlik}{1996}]{trepan}
Craven, M., and Shavlik, J.~W.
\newblock 1996.
\newblock Extracting tree-structured representations of trained networks.
\newblock In Touretzky, D.~S.; Mozer, M.~C.; and Hasselmo, M.~E., eds., {\em
  Advances in Neural Information Processing Systems 8}. MIT Press.
\newblock  24--30.

\bibitem[\protect\citeauthoryear{de Boves~Harrington}{2015}]{svm_dt}
de~Boves~Harrington, P.
\newblock 2015.
\newblock Support vector machine classification trees.
\newblock {\em Analytical chemistry} 87 21:11065--71.

\bibitem[\protect\citeauthoryear{Friedman}{2001}]{gbdt_with_multiplier}
Friedman, J.~H.
\newblock 2001.
\newblock Greedy function approximation: A gradient boosting machine.
\newblock {\em Ann. Statist.} 29(5):1189--1232.

\bibitem[\protect\citeauthoryear{Friedman}{2002}]{gradboost}
Friedman, J.~H.
\newblock 2002.
\newblock Stochastic gradient boosting.
\newblock {\em Comput. Stat. Data Anal.} 38(4):367--378.

\bibitem[\protect\citeauthoryear{Gatys, Ecker, and Bethge}{2016}]{ArtStyle}
Gatys, L.; Ecker, A.; and Bethge, M.
\newblock 2016.
\newblock A neural algorithm of artistic style.
\newblock {\em Journal of Vision} 16(12).

\bibitem[\protect\citeauthoryear{{Goodfellow}, {Shlens}, and
  {Szegedy}}{2014}]{adversarial}
{Goodfellow}, I.~J.; {Shlens}, J.; and {Szegedy}, C.
\newblock 2014.
\newblock {Explaining and Harnessing Adversarial Examples}.
\newblock {\em ArXiv e-prints}.

\bibitem[\protect\citeauthoryear{Guidotti \bgroup et al\mbox.\egroup
  }{2018a}]{LORE}
Guidotti, R.; Monreale, A.; Ruggieri, S.; Pedreschi, D.; Turini, F.; and
  Giannotti, F.
\newblock 2018a.
\newblock Local rule-based explanations of black box decision systems.
\newblock {\em CoRR} abs/1805.10820.

\bibitem[\protect\citeauthoryear{Guidotti \bgroup et al\mbox.\egroup
  }{2018b}]{survey_guidotti}
Guidotti, R.; Monreale, A.; Ruggieri, S.; Turini, F.; Giannotti, F.; and
  Pedreschi, D.
\newblock 2018b.
\newblock A survey of methods for explaining black box models.
\newblock {\em ACM Comput. Surv.} 51(5):93:1--93:42.

\bibitem[\protect\citeauthoryear{{Hinton}, {Vinyals}, and
  {Dean}}{2015}]{distillation2015}
{Hinton}, G.; {Vinyals}, O.; and {Dean}, J.
\newblock 2015.
\newblock {Distilling the Knowledge in a Neural Network}.
\newblock {\em ArXiv e-prints}.

\bibitem[\protect\citeauthoryear{Kim, Rudin, and Shah}{2014}]{bcm}
Kim, B.; Rudin, C.; and Shah, J.~A.
\newblock 2014.
\newblock The bayesian case model: A generative approach for case-based
  reasoning and prototype classification.
\newblock In Ghahramani, Z.; Welling, M.; Cortes, C.; Lawrence, N.~D.; and
  Weinberger, K.~Q., eds., {\em Advances in Neural Information Processing
  Systems 27}. Curran Associates, Inc.
\newblock  1952--1960.

\bibitem[\protect\citeauthoryear{Kindermans \bgroup et al\mbox.\egroup
  }{2017}]{Kindermans:2017}
Kindermans, P.-J.; Hooker, S.; Adebayo, J.; Alber, M.; Sch{\"u}tt, K.~T.;
  D{\"a}hne, S.; Erhan, D.; and Kim, B.
\newblock 2017.
\newblock The (un)reliability of saliency methods.
\newblock {\em CoRR} abs/1711.00867.

\bibitem[\protect\citeauthoryear{Kingma and Ba}{2014}]{adam}
Kingma, D.~P., and Ba, J.
\newblock 2014.
\newblock Adam: {A} method for stochastic optimization.
\newblock {\em CoRR} abs/1412.6980.

\bibitem[\protect\citeauthoryear{Lakkaraju \bgroup et al\mbox.\egroup
  }{2017}]{beta}
Lakkaraju, H.; Kamar, E.; Caruana, R.; and Leskovec, J.
\newblock 2017.
\newblock Interpretable {\&} explorable approximations of black box models.
\newblock {\em CoRR} abs/1707.01154.

\bibitem[\protect\citeauthoryear{LeCun and Cortes}{2010}]{mnist}
LeCun, Y., and Cortes, C.
\newblock 2010.
\newblock {MNIST} handwritten digit database.

\bibitem[\protect\citeauthoryear{Li \bgroup et al\mbox.\egroup
  }{2018}]{prototypes}
Li, O.; Liu, H.; Chen, C.; and Rudin, C.
\newblock 2018.
\newblock Deep learning for case-based reasoning through prototypes: A neural
  network that explains its predictions.
\newblock In {\em AAAI Conference on Artificial Intelligence}.

\bibitem[\protect\citeauthoryear{Lipton}{2018}]{lipton}
Lipton, Z.~C.
\newblock 2018.
\newblock The mythos of model interpretability.
\newblock {\em Queue} 16(3):30:31--30:57.

\bibitem[\protect\citeauthoryear{Lu, Setiono, and Liu}{1995}]{neurorule}
Lu, H.; Setiono, R.; and Liu, H.
\newblock 1995.
\newblock Neurorule: A connectionist approach to data mining.
\newblock In {\em Proceedings of the 21th International Conference on Very
  Large Data Bases}, VLDB '95,  478--489.
\newblock San Francisco, CA, USA: Morgan Kaufmann Publishers Inc.

\bibitem[\protect\citeauthoryear{Maas \bgroup et al\mbox.\egroup }{2011}]{imdb}
Maas, A.~L.; Daly, R.~E.; Pham, P.~T.; Huang, D.; Ng, A.~Y.; and Potts, C.
\newblock 2011.
\newblock Learning word vectors for sentiment analysis.
\newblock In {\em Proceedings of the 49th Annual Meeting of the Association for
  Computational Linguistics: Human Language Technologies},  142--150.
\newblock Portland, Oregon, USA: Association for Computational Linguistics.

\bibitem[\protect\citeauthoryear{Morcos \bgroup et al\mbox.\egroup
  }{2018}]{SingleDirGener}
Morcos, A.~S.; Barrett, D.~G.; Rabinowitz, N.~C.; and Botvinick, M.
\newblock 2018.
\newblock On the importance of single directions for generalization.
\newblock In {\em International Conference on Learning Representations}.

\bibitem[\protect\citeauthoryear{Morcos, Raghu, and
  Bengio}{2018}]{InsightCanonCorr}
Morcos, A.~S.; Raghu, M.; and Bengio, S.
\newblock 2018.
\newblock Insights on representational similarity in neural networks with
  canonical correlation.
\newblock {\em CoRR} abs/1806.05759.

\bibitem[\protect\citeauthoryear{Murthy, Kasif, and
  Salzberg}{1994}]{oblique_dt}
Murthy, S.~K.; Kasif, S.; and Salzberg, S.
\newblock 1994.
\newblock A system for induction of oblique decision trees.
\newblock {\em J. Artif. Intell. Res.} 2:1--32.

\bibitem[\protect\citeauthoryear{Nugent and Cunningham}{2005}]{cbr_for_bb}
Nugent, C., and Cunningham, P.
\newblock 2005.
\newblock A case-based explanation system for black-box systems.
\newblock {\em Artificial Intelligence Review} 24(2):163--178.

\bibitem[\protect\citeauthoryear{Papernot and McDaniel}{2018}]{DeepkNN}
Papernot, N., and McDaniel, P.
\newblock 2018.
\newblock Deep k-nearest neighbors: Towards confident, interpretable and robust
  deep learning.

\bibitem[\protect\citeauthoryear{Paszke \bgroup et al\mbox.\egroup
  }{2017}]{pytorch}
Paszke, A.; Gross, S.; Chintala, S.; Chanan, G.; Yang, E.; DeVito, Z.; Lin, Z.;
  Desmaison, A.; Antiga, L.; and Lerer, A.
\newblock 2017.
\newblock Automatic differentiation in pytorch.

\bibitem[\protect\citeauthoryear{Pedregosa \bgroup et al\mbox.\egroup
  }{2011}]{scikit-learn}
Pedregosa, F.; Varoquaux, G.; Gramfort, A.; Michel, V.; Thirion, B.; Grisel,
  O.; Blondel, M.; Prettenhofer, P.; Weiss, R.; Dubourg, V.; Vanderplas, J.;
  Passos, A.; Cournapeau, D.; Brucher, M.; Perrot, M.; and Duchesnay, E.
\newblock 2011.
\newblock Scikit-learn: Machine learning in {P}ython.
\newblock {\em Journal of Machine Learning Research} 12:2825--2830.

\bibitem[\protect\citeauthoryear{Ribeiro, Singh, and Guestrin}{2016}]{lime}
Ribeiro, M.~T.; Singh, S.; and Guestrin, C.
\newblock 2016.
\newblock "why should {I} trust you?": Explaining the predictions of any
  classifier.
\newblock {\em CoRR} abs/1602.04938.

\bibitem[\protect\citeauthoryear{Tan \bgroup et al\mbox.\egroup
  }{2017}]{distilltransparentaudit}
Tan, S.; Caruana, R.; Hooker, G.; and Lou, Y.
\newblock 2017.
\newblock Distill-and-compare: Auditing black-box models using transparent
  model distillation.

\bibitem[\protect\citeauthoryear{Tan \bgroup et al\mbox.\egroup
  }{2018}]{distilltransparent}
Tan, S.; Caruana, R.; Hooker, G.; and Gordo, A.
\newblock 2018.
\newblock Transparent model distillation.

\bibitem[\protect\citeauthoryear{Zeiler and Fergus}{2014}]{VisConvNet}
Zeiler, M.~D., and Fergus, R.
\newblock 2014.
\newblock Visualizing and understanding convolutional networks.
\newblock In Fleet, D.; Pajdla, T.; Schiele, B.; and Tuytelaars, T., eds., {\em
  Computer Vision -- ECCV 2014},  818--833.
\newblock Cham: Springer International Publishing.

\bibitem[\protect\citeauthoryear{Zhang \bgroup et al\mbox.\egroup
  }{2018}]{DeepFeatures}
Zhang, R.; Isola, P.; Efros, A.~A.; Shechtman, E.; and Wang, O.
\newblock 2018.
\newblock The unreasonable effectiveness of deep features as a perceptual
  metric.
\newblock In {\em The IEEE Conference on Computer Vision and Pattern
  Recognition (CVPR)}.

\bibitem[\protect\citeauthoryear{Zhang, Nian~Wu, and
  Zhu}{2018}]{interpretable_cnn}
Zhang, Q.; Nian~Wu, Y.; and Zhu, S.-C.
\newblock 2018.
\newblock Interpretable convolutional neural networks.
\newblock In {\em The IEEE Conference on Computer Vision and Pattern
  Recognition (CVPR)}.

\end{thebibliography}

\end{document}